\documentclass[runningheads]{llncs}
\usepackage{graphicx}
\usepackage{amsmath,amssymb} 
\usepackage{color}
\usepackage{enumerate}
\usepackage{multirow}
\usepackage{hyperref}
\usepackage[nobiblatex]{xurl}
\begin{document}
\pagestyle{headings}
\mainmatter

\newcommand{\ie}{\textit{i}.\textit{e}., }
\newcommand{\eg}{\textit{e}.\textit{g}., } 
\newcommand{\et}{\textit{et al.}}
%
%
%

\title{Spatio-channel Attention Blocks \\for Cross-modal Crowd Counting}


\author{Youjia Zhang\orcidID{0000-0002-5671-3232} \and
Soyun Choi\orcidID{0000-0002-2701-4782} \and
Sungeun Hong\orcidID{0000-0003-1774-9168}}
\authorrunning{Y. Zhang et al.}
%

\institute{Department of  Electrical  and Computer Engineering, 
Inha University, South Korea
\email{\{zhangyoujia,sychoi\}@inha.edu}, {csehong@inha.ac.kr}}

\maketitle              
\begin{abstract}
Crowd counting research has made significant advancements in real-world applications, but it remains a formidable challenge in cross-modal settings.
Most existing methods rely solely on the optical features of RGB images, ignoring the feasibility of other modalities such as thermal and depth images. 
The inherently significant differences between the different modalities and the diversity of design choices for model architectures make cross-modal crowd counting more challenging.
In this paper, we propose Cross-modal Spatio-Channel Attention (CSCA) blocks, which can be easily integrated into any modality-specific architecture.
The CSCA blocks first spatially capture global functional correlations among multi-modality with less overhead through spatial-wise cross-modal attention.
Cross-modal features with spatial attention are subsequently refined through adaptive channel-wise feature aggregation.
In our experiments, the proposed block consistently shows significant performance improvement across various backbone networks, resulting in state-of-the-art results in RGB-T and RGB-D crowd counting. Code is available at \textcolor{red}{\url{https://github.com/VCLLab/CSCA}}.

\end{abstract}
\section{Introduction}

Crowd counting is a core and challenging task in computer vision, which aims to automatically estimate the number of pedestrians in unconstrained scenes without any prior knowledge. Driven by real-world applications, including traffic monitoring \cite{ali2019leveraging}, social distancing monitoring \cite{ghodgaonkar2020analyzing}, and other security-related scenarios \cite{xu2017efficient}, crowd counting plays an indispensable role in various fields.
Over the last few decades, an increasing number of researchers have focused on this topic, achieving significant progress \cite{wan2021generalized,zhang2017fcn,zhang2017visual,walach2016learning} in counting tasks.

In general, the various approaches for crowd counting can be roughly divided into two categories: detection-based methods and regression-based methods. Early works \cite{dalal2005histograms,leibe2005pedestrian,enzweiler2008monocular,tuzel2008pedestrian,felzenszwalb2010object,wu2007detection} on crowd counting mainly use detection-based approaches, which detect human heads in an image as can be seen in drone object detection tasks \cite{du2019visdrone,hong2019patch}. In fact, most detection-based methods assume that the crowd in the image is made up of individuals that can be detected by the given detectors. There are still some significant limitations for tiny heads in higher density scenarios with overlapping crowds and serious occlusions. To reduce the above problems, some researchers \cite{chan2008privacy,chan2009bayesian,idrees2013multi,shao2016slicing} introduce regression-based methods where they view crowd counting as a mapping problem from an image to the count or the crowd-density map. In this way, the detection problem can be circumvented.
Recent works \cite{zhang2016single,li2018csrnet,zhang2019relational,sindagi2017generating} have shown the significant success of density map estimation in RGB crowd counting, and it has become the mainstream method for the counting task in complex crowd scenes.

\begin{figure}[t]
\begin{center}
\setlength{\abovecaptionskip}{0.cm}
 \vspace{-0.3cm}
\def\arraystretch{0.5}
\begin{tabular}{@{}c@{}c@{}c@{}c}

\includegraphics[width=0.22\linewidth]{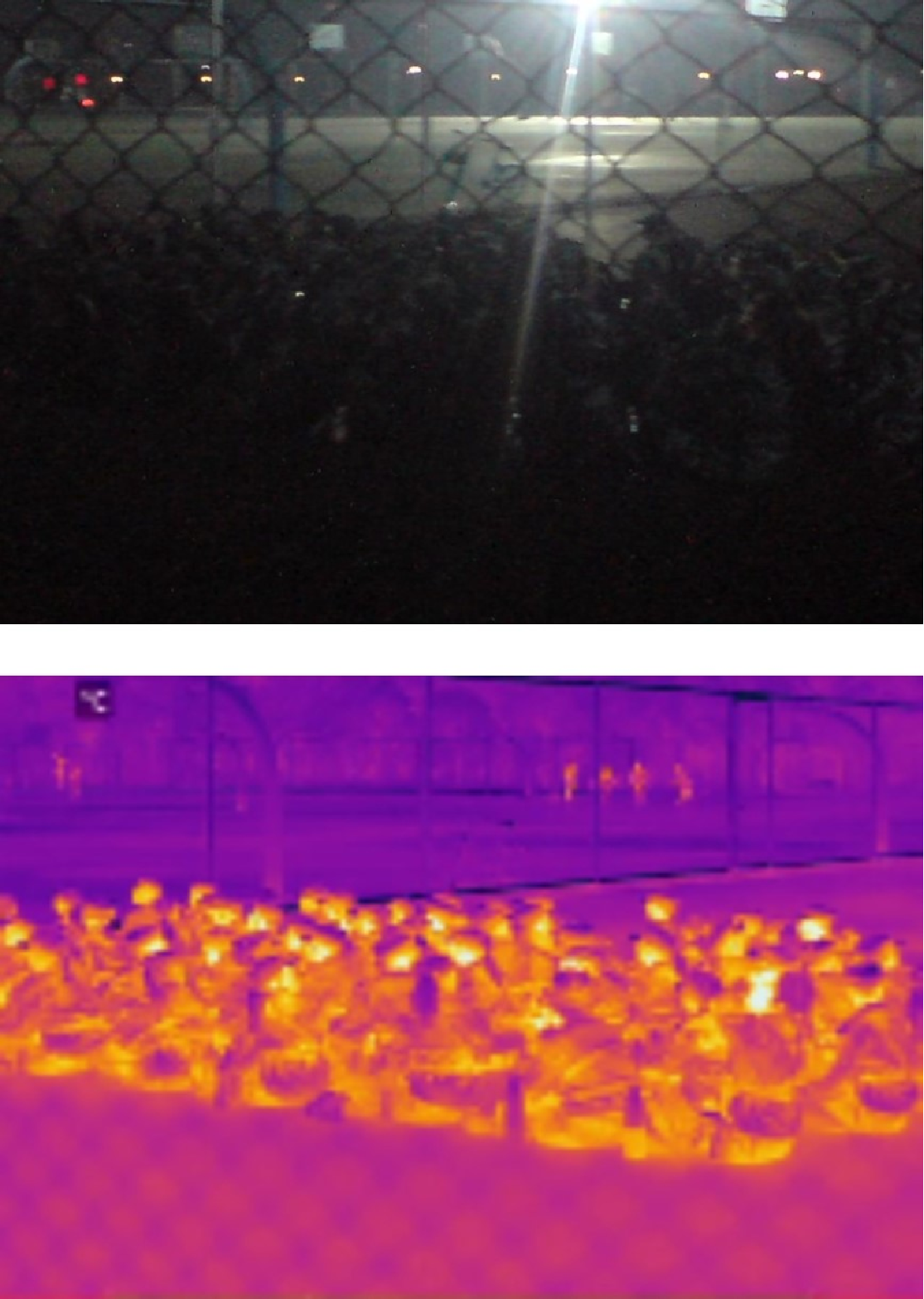} & \:\:\:
\includegraphics[width=0.22\linewidth]{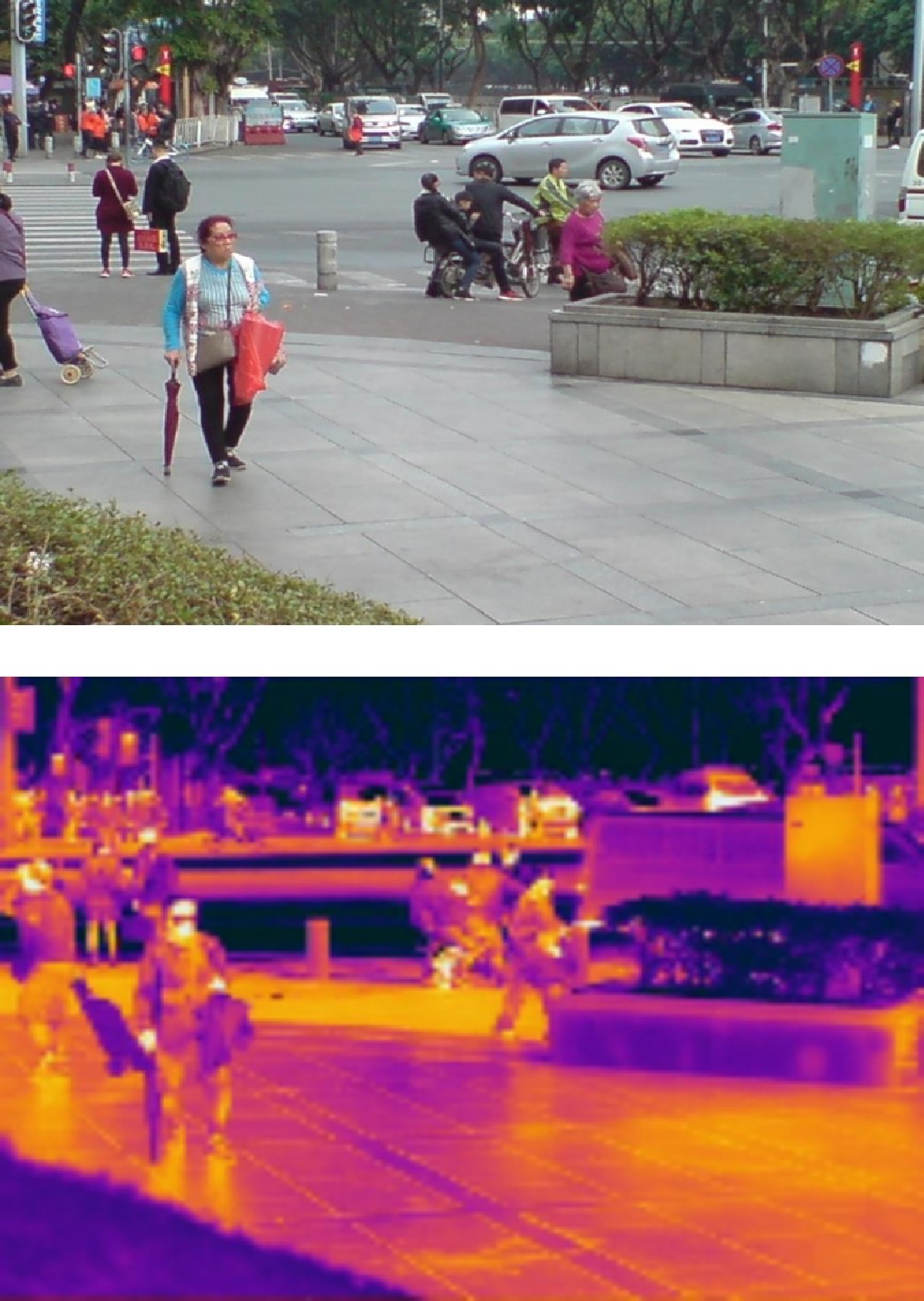} & \:\:\:
\includegraphics[width=0.22\linewidth]{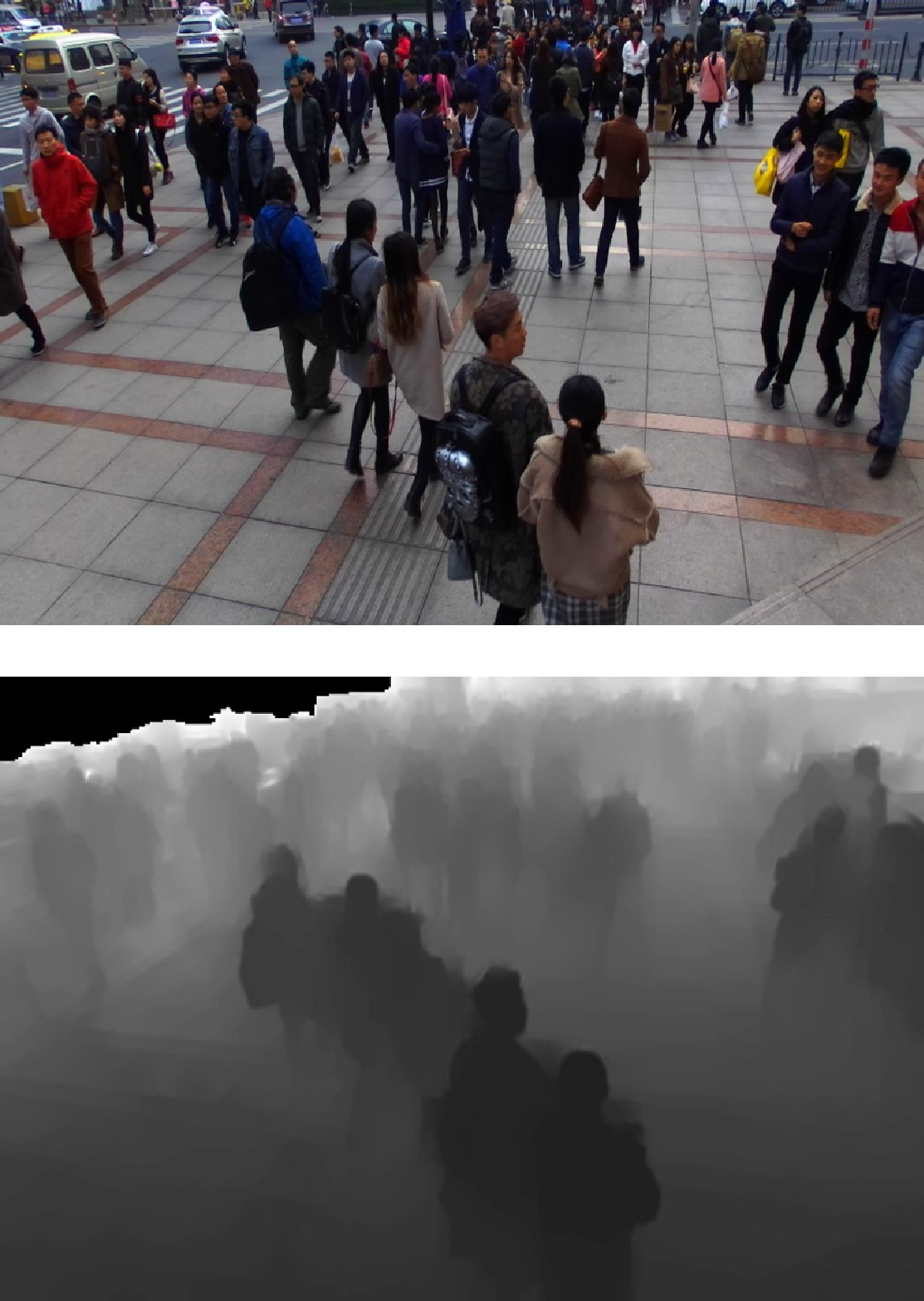} & \:\:\:
\includegraphics[width=0.22\linewidth]{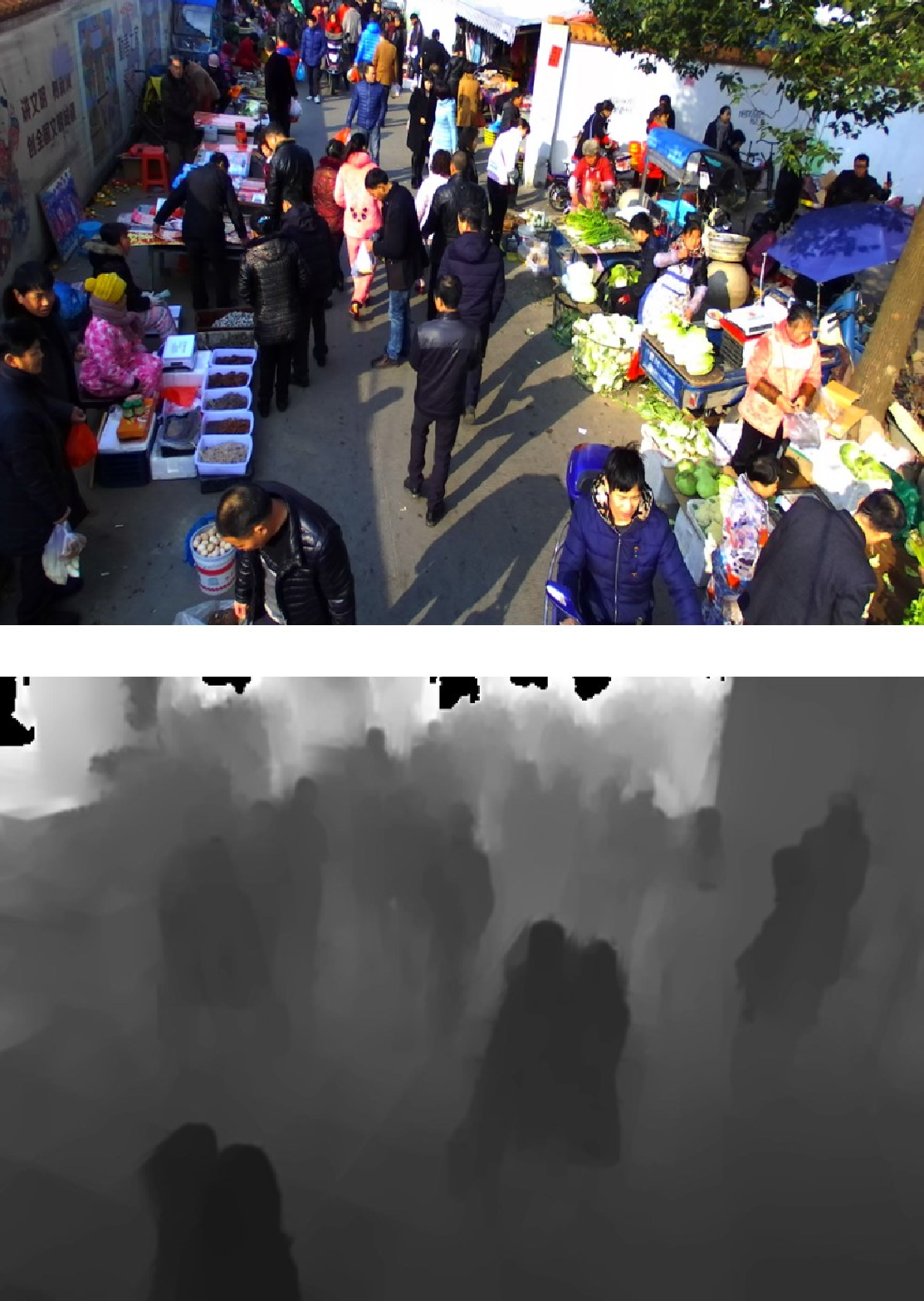} \\\\
\scriptsize
{(a) RGB-T (RGB$<$T)} & \:\:\:  \scriptsize {(b) RGB-T (RGB$>$T)} &\:\:\: \scriptsize{(c) RGB-D (RGB$<$D)} &\:\:\: \scriptsize{(d) RGB-D (RGB$>$D)}
\end{tabular}
\end{center}
\caption{
Visualization of RGB-T and RGB-D pairs. (a) shows the positive effects of thermal images in extremely poor illumination. (b) shows the negative effects caused by additional heating objects. In case (c), the depth image provides additional information about the position and size of the head, but in scene (d), part of the head information in the depth image is corrupted by cluttered background noise.}
\label{Fig:1}
 \vspace{-0.3cm}
\end{figure}

Despite substantial progress, crowd counting remains a long-standing challenging task for various situations such as occlusion, high clutter, non-uniform lighting, non-uniform distribution, etc. Due to the limitations of methods or insufficient data sources, most previous works have focused on the RGB crowd counting and failed to obtain high-quality density maps in unconstrained scenarios. 
RGB images can provide strong evidence in bright illumination conditions while they are almost invisible in the dark environment. Recently, some researchers \cite{liu2021cross,liu2021visdrone} indicate that the thermal feature helps better distinguish various targets in poor illumination conditions. Therefore, as shown in Fig.~\ref{Fig:1}-(a), thermal images can be used to improve the performance of crowd counting tasks. However, from Fig.~\ref{Fig:1}-(b), we can find that additional heating objects seriously interfere with the thermal image. 

Furthermore, besides exploiting various contextual information from the optical cues, the depth feature \cite{bondi2014real,xu2019depth,lian2019density} has recently been utilized as supplementary information of the RGB feature to generate improved crowd density maps. From Fig.~\ref{Fig:1}-(c), we observe that crowd movement leads to aggregation behavior and large variation in head size. Thus, it is insufficient to generate pixel-wise density maps using only RGB data. Depth data naturally complements RGB information by providing the 3D geometry to 2D visual information, which is robust to varying illumination and provides additional information about the localization and size of heads. However, as shown in Fig.~\ref{Fig:1}-(d),the depth information would be critically affected by noise under the condition of cluttered background.

Despite the many advantages of using additional modalities besides RGB in crowd counting, there are few related studies in crowd counting.
In this study, we focus on two main issues that need to be addressed for a more robust cross-modal crowd counting as follows:

\begin{enumerate}[i]
\item 
Crowd counting data is obtained from the different modalities, which means that they vary significantly in distribution and properties. Since there are substantial variations between multimodal data, how to effectively identify their differences and unify multiple types of information is still an open problem in cross-modal crowd counting. Additionally, since additional modality other than RGB is not fixed (either thermal or depth), it is very critical to adaptively capture intra-modal and inter-modal correlation.
 
\item 
For cross-modal crowd counting, various types can be given as additional modalities besides RGB. 
At this time, constructing a cross-modal model architecture from scratch usually entails a lot of trial and error.
We have seen the well-explored model architectures show promising results in various fields through transfer learning or model tuning.
In this respect, if the widely used RGB-based unimodal architecture can be extended to multimodal architecture without bells and whistles, we can relieve the effort of design choice of the model architecture.
\end{enumerate}

In this paper, we propose a novel plug-and-play module called Cross-modal Spatio-Channel Attention (CSCA) block consisting of two main modules.
First, Spatial-wise Cross-modal Attention (SCA) module utilizes an attention mechanism based on the triplet of `Query', `Key', and `Value' widely used in non-local-based models \cite{wang2018non,zhu2019asymmetric,yang2020weakly}.
Notably, conventional non-local blocks are unsuitable for cross-modal tasks due to their high computational complexity and the fact that they are designed to exploit self-attention within a single modality.
To address these issues, SCA module is deliberately designed to understand the correlation between cross-modal features, and reduces computational overhead through the proposed feature re-assembling scheme.
Given an input feature map with size C×H×W, the computational complexity of a typical non-local block is $O(CH^2 W^2)$ while our SCA block can reduce the complexity by a factor of $G_{l}$ to $O(CH^2 W^2/G_{l})$. Here, the $G_l$ refers to the spatial re-assembling factor in the $l^{th}$ block of the backbone network.

Importantly, we note that non-local blocks widely used in various vision tasks focus on spatial correlation of features but do not directly perform feature recalibration at the channel level.
To supplement the characteristics of non-local-based blocks considering only spatial correlation, our Channel-wise Feature Aggregation (CFA) module adaptively adjusts the spatially-correlated features at the channel level.
Unlike previous works, which use pyramid pooling feature differences to measure multimodal feature variation \cite{liu2021cross} or fine-tune the density map \cite{li2022rgb}, our SCA aims to spatially capture global feature correlations between multimodal features to alleviate misalignment in multimodal data.
Moreover, instead of adding auxiliary tasks for other modal information \cite{bondi2014real,xu2019depth,lian2019density}, our method dynamically aggregate complementary features along the channel dimension by the CFA block.
We show that the proposed two main modules complement each other on two challenging RGB-T \cite{liu2021cross} and RGB-D \cite{lian2019density}. Additionally,  we conduct various ablation studies and comparative experiments in cross-modal settings that have not been well explored.
In summary, the major contributions of this work are as follows:

\begin{itemize}
\item We propose a new plug-and-play module that can significantly improve performance when integrated with any RGB-based crowd counting model.
\end{itemize}

\begin{itemize}
\item Our cross-modal attention with feature re-assembling is computationally efficient and has higher performance than conventional non-local attention. Moreover, our channel-level recalibration further improves the performance.
\end{itemize}

\begin{itemize}
\item The proposed CSCA block consistently demonstrates significant performance enhancements across various backbone networks and shows state-of-the-art performance in RGB-T and RGB-D crowd counting.
\end{itemize}

\begin{itemize}
\item We extensively conduct and analyze comparative experiments between multi-modality and single-modality, which have not been well explored in existing crowd counting, and this can provide baselines for subsequent studies.
\end{itemize}

\section{Related Works}

\subsection{Detection-based methods}
Early works of crowd counting focus on detection-based approaches. The initial method \cite{lin2001estimation,viola2005detecting} mainly employs detectors based on hand-crafted features.  The typical traditional methods \cite{dalal2005histograms,leibe2005pedestrian,enzweiler2008monocular,li2008estimating} train classifiers to detect pedestrians using wavelet, HOG, edge, and other features extracted from the whole body, which degrades seriously for those very crowded scenes. With the increase of crowd density, the occlusion between people becomes more and more serious. Therefore, methods \cite{felzenszwalb2010object,wu2007detection} based on partial body detection (such as head, shoulders, etc.) are proposed to deal with the crowd counting problem. Compared with the overall detection, this method has a slight improvement in effect. Recently, detection-based \cite{lian2019density,liu2018decidenet,valencia2021vision,bathija2019visual} methods have also been further improved with the introduction of more advanced object detectors. For example, some works \cite{liu2018decidenet,lian2019density} design specific architectures such as Faster RCNN \cite{ren2015faster} or RetinaNet \cite{lin2017focal} to promote the performance of crowd counting by improving the robustness of detectors for tiny head detection. But, they still present unsatisfactory results on challenging images with very high densities, significant occlusions, and background clutter in extremely dense crowds, which are quite common in crowd counting tasks.

\subsection{Regression--based methods}
 \vspace{-1mm}
Some of the recent approaches \cite{chan2009bayesian,zhang2016single,li2018csrnet,fu2015fast,wang2015deep,boominathan2016crowdnet,sindagi2017cnn,liu2018leveraging} attempt to tackle this problem by bypassing completely the detection problem. They directly address it as a mapping problem from images to counts or crowd density maps. Recently, numerous models utilize basic CNN layers \cite{fu2015fast,wang2015deep}, multi-column based models \cite{zhang2016single,boominathan2016crowdnet}, or single-column based models \cite{li2018csrnet,ma2019bayesian,cao2018scale} to extract rich information for density estimation and crowd counting. To capture multi-scale information at different resolutions, MCNN \cite{zhang2016single} proposes a three branches architecture. And, CrowdNet \cite{boominathan2016crowdnet} combines low-level features and high-level semantic information. To reduce the information redundancy caused by the multi-column method, SANet \cite{cao2018scale} is built on the Inception architecture. And, CSRNet \cite{li2018csrnet} adopts dilated convolution layers to expand the receptive field. Regression-based methods capture low-level features and high-level semantic information through multi-scale operations or dilated convolution, which show remarkable success in crowd counting tasks. However, most methods only focus on the RGB images and may fail to accurately recognize the semantic objects in unconstrained scenarios.

\subsection{Multi-modal learning for crowd counting} 
Recently, multimodal feature fusion has been widely studied across various fields \cite{hong2018cbvmr,sun2019leveraging,nagrani2021attention}. The cross-modal feature fusion aims to learn better feature representations by exploiting the complementarity between multiple modalities. Some cross-modal fusion methods \cite{fu2020jl,sun2019leveraging} fuse the RGB source and the depth source in the “Early Fusion” or “Late Fusion” way. Besides,  inspired by the success of multi-task learning in various computer vision tasks, some researchers \cite{zhao2019leveraging,shi2019counting} utilize the depth feature as the auxiliary information to boost the performance on the RGB crowd counting task. Similarly, the perspective information \cite{shi2019revisiting} during the density estimation procedure is also treated as auxiliary data to improve the performance of RGB crowd counting.  However, these methods rely on handcrafted features instead of fusing multimodal data with predicting weights from different modalities. Moreover, some researchers \cite{lian2019density,fu2012real} combine the depth information into the detection procedure for the crowd counting task, which would be limited by the performance of the detectors. The IADM \cite{liu2021cross} proposes to collaboratively represent the RGB and thermal images. However, it does not fully consider the different characteristics and feature space misalignment of RGB data and thermal imaging data. Different from previous work, in this work, we propose a novel cross-modal spatio-channel attention block, which adaptively aggregates cross-modal features based on both spatial-wise and channel-wise attention. Furthermore, our SCA block is robust to cross-modal feature variation at lower complexity compared to the conventional non-local block.

\begin{figure*}[t]
\centering
 \setlength{\abovecaptionskip}{0.cm}
\vspace{-0.3cm}
\includegraphics[width=0.9\textwidth]{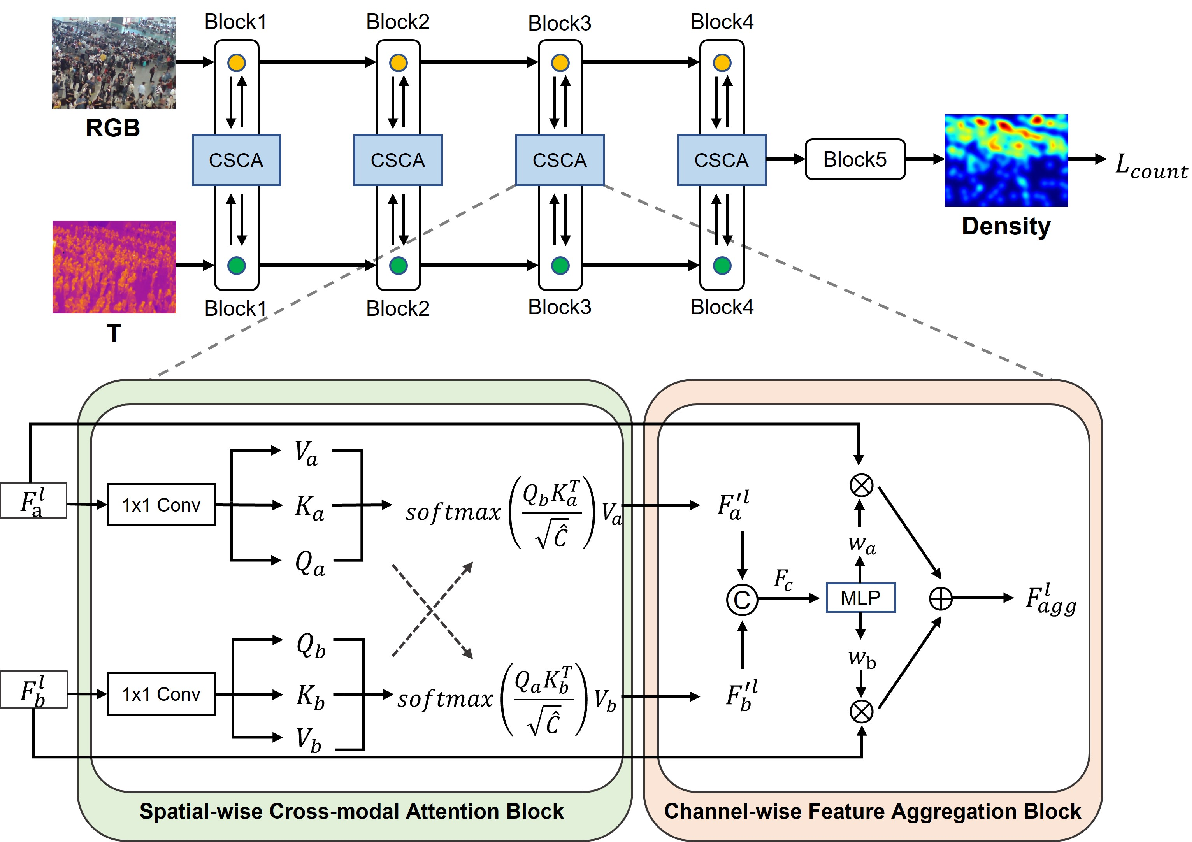} 
\caption{
The architecture of the proposed unified framework for extending existing baseline models from unimodal crowd counting to multimodal scenes. Our CSCA module is taken as the cross-modal solution to fully exploit the multimodal complementarities. Specifically, the CSCA consists of SCA to model global feature correlations among multimodal data, and CFA to dynamically aggregate complementary features.}
\label{Fig:2}
\vspace{-0.4cm}
\end{figure*}

\section{Proposed Method}
\subsection{Overview}
We proposed a unified framework to extend the existing baseline models from unimodal crowd counting to the multimodal scene.
As shown in Fig.~\ref{Fig:2}, the framework for cross-modal crowd counting consists of two parts: modality-specific branches and the Cross-modal Spatio-Channel Attention (CSCA) block. Given pairs of multimodal data like RGB-T or RGB-D pairs, we first utilize the backbone network (\eg BL \cite{ma2019bayesian}, CSRNet \cite{li2018csrnet}, etc.) to extract the modality-specific features of RGB images and thermal images. To fully exploit the multimodal complementarities, we embed CSCA after different convolutional blocks of the two modality-specific branches to further update the feature maps and get the complementary fusion information.

To ensure informative feature interaction between modalities, our CSCA consists of two operations: Spatial-wise Cross-modal Attention (SCA) block and Channel-wise Feature Aggregation (CFA) block, as illustrated in the bottom part of Fig.~\ref{Fig:2}. The SCA block utilizes an improved cross-modal attention block to model the global feature correlations between RGB images and thermal/depth images, which has a larger receptive field than the general CNN modules and facilitates complementary aggregation operation. And, the CFA block adaptively extracts and fuses cross-modal information based on channel correlations. Consequently, the framework can capture the non-local dependencies and aggregate the complementary information of cross-modal data to expand from unimodal crowd counting to multimodal crowd counting.

\begin{figure*}[t]
\centering
\setlength{\abovecaptionskip}{-0.1cm}
\vspace{-0.3cm}
\includegraphics[width=1\textwidth]{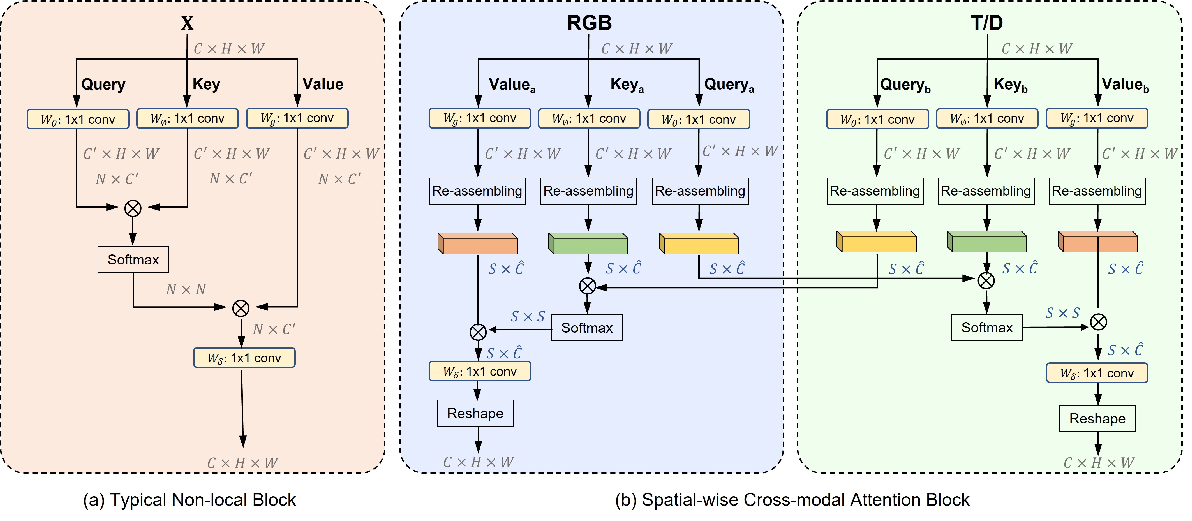} 
\caption{
Comparison between block structures (a)  typical non-local block
(b) proposed SCA block where the blue font indicates dimension transformation of SCA block, $\hat{C}=C^{'}G_l$ and $S\times{\hat{C}}=N\times{C^{'}}$ while $S=N/G_l$.
}
\label{Fig:3}
  \vspace{-3mm}
\end{figure*}

\subsection{Spatial-wise Cross-modal Attention (SCA)}

SCA module aims to capture global feature correlations among multimodal data with less overhead based on cross-modal attention. Recently, the Transformer-based architecture \cite{vaswani2017attention} has achieved comparable or even superior performance on computer vision tasks. The multi-head attention as one of the most important parts of the Transformer consists of several non-local attention layers running in parallel. A typical non-local block \cite{wang2018non} is shown in Fig.~\ref{Fig:3}-(a), which aims to capture the global relationships among all locations of the input image. Consider an input feature map $X\in{\mathcal{R}^{C\times{H}\times{W}}}$, where $C$, $H$, and $W$ indicate the channel number, spatial height, and width, respectively. The feature map is first projected and flattened into three vectors: Query $Q\in{\mathcal{R}^{N\times{C^{'}}}}$, Key $K\in{\mathcal{R}^{N\times{C^{'}}}}$ and Value $V\in{\mathcal{R}^{N\times{C^{'}}}}$ by three different $1\times1$ convolutions, where $N=H\times{W}$, and $C^{'}$ refers to the new embedding channel, $C^{'}=C/2$. Then, the triplet of Query, Key, and Value are used to encode the relationship at any position in the feature map, thereby capturing the global context information. The output can be computed as a weighted sum of values where the weight is assigned to each value by the compatibility of the query with the corresponding key. This procedure can be computed efficiently with matrix multiplications as:

\begin{equation}
Z=f(QK^T)V, \label{1}
\end{equation}
where the $f(\cdot)$ is a normalization function, such as softmax, rescaling, etc. Finally, the non-local block recovers the channel dimension from $C^{'}$ to C by an additional $1\times1$ convolution layer.

However, the typical non-local block can only work on unimodal data, and the high complexity introduced by matrix multiplication hinders its use. To alleviate these problems, we propose the SCA block. Different from the typical non-local block, the input of our SCA Block is a pair of images with the different modalities. And it's Key and Value are from the same modality, while the Query is from another modality. For example, the Key and Value are from the RGB image modality, and the Query is generated from thermal image or depth image modality, and vice versa.

\begin{equation}
Z_{a}=softmax(\frac{Q_{b}K_{a}^T}{\sqrt{\hat{C}}})V_a,\label{2}
\end{equation}
where the $a$ means the RGB modality and $b$ means another modality (\ie thermal or depth modality). The output of the SCA block (Fig.~\ref{Fig:3}-(b)), is calculated by multiplying Value with attention weights, which presents the similarity between the corresponding Query in modality $b$ and all the Keys in another modality $a$. In other words, the SCA block aggregates and aligns the information from two modalities. 

Thanks to the great power of cross-attention, many works \cite{zhu2019asymmetric,mehta2021mobilevit,li2021trear,xu2021cdtrans} have utilized it to solve the misalignment. However, the matrix multiplication operation is very time and memory-consuming compared to convolutions and activation blocks. Some researchers \cite{liu2021cross} alleviate the problem by reducing the dimensions of the Key and Value through the pooling operation. But, it is hard to ensure that the performance would not degenerate too much by losing a lot of global information. Instead of the pooling function, we perform re-assembling operations in the SCA block to randomly divide the feature map into $G_l$ patches in the spatial dimension, where $l$ means the $l^{th}$ block of the backbone network. Thus, the Key, Value, and Query vectors in our SCA block become $K$,$V$,$Q\in{\mathcal{R}^{S\times{\hat{C}}}}$, where $S=HW/G_l$, $\hat{C}=C^{'}G_l$. Finally, we recover the channel dimension and reshape the result to $C\times{H}\times{W}$ size. Therefore, the computational cost of our SCA block is $O(CH^2 W^2/G_l)$, which is $G_l$ times less than the computational complexity $O(CH^2 W^2)$ of a typical non-local block.

\subsection{Channel-wise Feature Aggregation (CFA)}
The SCA block can capture global features at all locations in multimodal images by spatial-wise attention. To make full use of the complementarity of multimodal data, we need to complementarily aggregate cross-modal features according to their characterization capabilities. To achieve this, we propose a CFA block to fusion the cross-modal features and capture channel-wise interactions. As shown in the bottom part of Fig.~\ref{Fig:2}, we first concatenate the outputs of the SCA block to get the $F_c$. We then use an MLP layer and softmax function to learn the weight vectors  $w_a,w_b\in{\mathcal{R}^{C\times{H}\times{W}}}$, which are used to re-weight the RGB feature and thermal/depth feature across the channels.

\begin{equation}
[w_a,w_b]=softmax(MLP(F_c)). \label{3}
\end{equation}

To fully exploit the aggregated information and suppress feature noise and redundancy simultaneously, we fusion the cross-modal features using the two channel-wise adaptive weights. The final aggregation feature map $F^l_{agg}$ at $l^{th}$ block can be obtained by the following formula:

\begin{equation}
F^l_{agg}=w_a*F^l_{a}+w_b*F^l_b. \label{4}
\end{equation}

Through the CFA block, the network can increase sensitivity to informative features on the important channels and enhance the final feature map. In practice, for each layer $l$, the updated feature maps $f^l_{a}= (F^l_{agg}+F^l_{a})/2$ and $f_b^l= (F^l_{agg}+F^l_b)/2$ are propagated to the next blocks except the last layer. Hence, the module can adaptively provide more discriminative and richer characteristics to increase cross-modal crowd counting performance.

\section{Experiments}
To evaluate our method, we carry out detailed experiments on two cross-modal crowd counting datasets: RGBT-CC benchmark \cite{liu2021cross} and ShanghaiTechRGBD \cite{lian2019density}. And we also provide a thorough ablation study to validate the key components of our approach. Experimental results show that the proposed method considerably improves the performance of the cross-modal crowd counting.

\subsection{Implementation details}

In this work, our code is based on PyTorch \cite{paszke2019pytorch}. Following \cite{liu2021cross}, our model is implemented with three backbone networks: BL \cite{ma2019bayesian}, MCNN \cite{zhang2016single}, and CSRNet \cite{li2018csrnet}. To make fair comparisons, we maintain a similar number of parameters to the original backbone models, and the same channel setting with the work \cite{liu2021cross}, which takes $70\%$, $60\%$, and $60\%$ of the original values.

For the input size, we feed the $640\times480$ RGB-T pairs and $1920\times1080$ RGB-D pairs, respectively. Notably, the original loss function of the adopted backbone network is used to train our framework. During training, our framework is optimized by Adam \cite{kingma2014adam} with a 1e-5 learning rate. All the experiments are conducted using $1\times$ RTX A6000 GPU.

\textbf{\textit{Datasets:}}
The popular datasets, RGBT-CC \cite{liu2021cross} and ShanghaiTechRGBD \cite{lian2019density}, in the cross-modal crowd counting field are used in this study. The RGBT-CC dataset is a new challenging benchmark for RGB-T crowd counting, including 2,030 RGB-thermal pairs with resolution $640\times480$. All the images are divided into three sets: training set with 1030 pairs, validation set with 200 pairs, and testing set with 800 pairs, respectively. The ShanghaiTechRGBD dataset is a large-scale RGB-D dataset, which consists of 1193 training RGB-D image pairs and 1000 test RGB-D image pairs with a fixed resolution of $1920\times1080$. 

\textbf{\textit{Evaluation Metrics:}}
As commonly used in previous works, we adopt the Root Mean Square Error (RMSE) and the Grid Average Mean absolute Error (GAME) as the evaluation metrics.
\begin{equation}
\begin{aligned}
RMSE=\sqrt{\frac{1}{N}\sum_{i=1}^{N}{(P_i-\hat{P_i})}^2} \\
GAME(L)=\frac{1}{N}\sum_{i=1}^{N}\sum_{l=1}^{4^L}|P^l_i-\hat{P}^l_i|,
\end{aligned}
\label{5}
\end{equation}
where N means numbers of testing images, $P_i$ and $\hat{P_i}$ refer to the total count of the estimated density map and corresponding ground truth, respectively. Different from the RMSE, the GAME divides each image into $4^L$ non-overlapping regions to measure the counting error in each equal area, where $P^l_i$ and $\hat{P}^l_i$ are the estimated count and ground truth in a region $l$ of image $i$. Note that $GAME(0)$ is equivalent to Mean Absolute Error (MAE).

\subsection{Effectiveness of plug-and-play CSCA blocks}

To evaluate the effectiveness of the proposed plug-and-play CSCA module, we easily incorporate CSCA as a plug-and-play module into backbone networks (\eg MCNN, CSRNet, and BL), and extend existing models from unimodal to multimodal on the RGBT-CC dataset. We report the result in Table~\ref{table:1}. Note that for all baseline methods in the table, we estimate crowd counts by feeding the concatenation of RGB and thermal images. As can be observed, almost all instances of CSCA based method significantly outperform the corresponding backbone networks. Especially, our method far outperforms MCNN by 4.52 in MAE and 10.60 in RMSE. Compared to the CSRNet backbone, our CSCA provides a 3.88 and 4.17 improvement in MAE and RMSE, respectively. For the BL backbone, our method surpasses the baseline on all evaluation metrics.

Similarly, we also apply CSCA to the RGB-D crowd counting task. Furthermore, previous research \cite{liu2021cross} showed that MCNN is inefficient on the ShanghaiTechRGBD dataset with high-resolution images due to its time-consuming multi-column structure. So here we only compare the two backbone networks: BL and CSRNet. As shown in Table~\ref{table:2}, our framework consistently outperforms their corresponding backbone networks on all evaluation metrics. In conclusion, all experimental results confirm that CSCA is universal and effective for cross-modal crowd counting.

\setlength{\tabcolsep}{4pt}
\begin{table}[t]
\begin{center}
\caption{The performance of our model implemented with different backbones on RGBT-CC.
}
\label{table:1}
\resizebox{0.89\columnwidth}{!}{
\begin{tabular}{cccccc}
\hline
Backbone &  GAME(0)$\downarrow$ & GAME(1)$\downarrow$ & GAME(2)$\downarrow$ & GAME(3)$\downarrow$ & RMSE$\downarrow$\\
\hline
MCNN\cite{zhang2016single} & 21.89 & 25.70 & \textbf{30.22} & \textbf{37.19} & 37.44\\
MCNN+CSCA & \textbf{17.37} & \textbf{24.21} & 30.37 & 37.57 & \textbf{26.84} \\
\hline
CSRNet\cite{li2018csrnet} & 20.40 & 23.58 & \textbf{28.03} & \textbf{35.51} & 35.26\\
CSRNet+CSCA & \textbf{17.02} &\textbf{23.30} & 29.87 & 38.39 & \textbf{31.09} \\
\hline
\noalign{\smallskip}
BL\cite{ma2019bayesian} & 18.70 & 22.55 & 26.83 & 34.62 & 32.64\\
BL+CSCA & \textbf{14.32} & \textbf{18.91} & \textbf{23.81} & \textbf{32.47} & \textbf{26.01}\\
\hline
\end{tabular}
}
\end{center}
\end{table}
\setlength{\tabcolsep}{4pt}

\setlength{\tabcolsep}{4pt}
\begin{table}[t]
\begin{center}
\caption{The performance of our model implemented with different backbones on ShanghaiTechRGBD.
}
\label{table:2}
\resizebox{0.89\columnwidth}{!}{
\begin{tabular}{cccccc}
\hline
Backbone &  GAME(0)$\downarrow$ & GAME(1)$\downarrow$ & GAME(2)$\downarrow$ & GAME(3)$\downarrow$ & RMSE$\downarrow$\\
\hline
\noalign{\smallskip}
BL\cite{ma2019bayesian} & 8.94 & 11.57 & 15.68 & 22.49 & 12.49\\
BL+CSCA & \textbf{5.68} & \textbf{7.70} & \textbf{10.45} & \textbf{15.88} & \textbf{8.66}\\
\hline
CSRNet\cite{li2018csrnet} & 4.92 & 6.78 & 9.47 & 13.06 & 7.41\\
CSRNet+CSCA & \textbf{4.39} & \textbf{6.47} & \textbf{8.82} & \textbf{11.76} & \textbf{6.39} \\
\hline
\end{tabular}
}
\end{center}
\vspace{-0.5cm}
\end{table}
\setlength{\tabcolsep}{4pt}

\setlength{\tabcolsep}{4pt}
\begin{table}[t]
\begin{center}
\caption{Counting performance evaluation on RGBT-CC.
}
\label{table:3}
\resizebox{0.89\columnwidth}{!}{
\begin{tabular}{cccccc}
\hline
Backbone &  GAME(0)$\downarrow$ & GAME(1)$\downarrow$ & GAME(2)$\downarrow$ & GAME(3)$\downarrow$ & RMSE$\downarrow$\\
\hline
\noalign{\smallskip}
UCNet \cite{zhang2020uc} & 33.96 & 42.42 & 53.06 & 65.07 & 56.31\\
HDFNet \cite{pang2020hierarchical} & 22.36 & 27.79 & 33.68 & 42.48 & 33.93\\
BBSNet \cite{fan2020bbs} & 19.56 & 25.07 & 31.25 & 39.24 & 32.48\\
MVMS \cite{zhang2019wide} & 19.97 & 25.10 & 31.02 & 38.91 & 33.97\\
MCNN+IADM \cite{liu2021cross} & 19.77 & 23.80 & 28.58 & 35.11 & 30.34\\
CSRNet+IADM \cite{liu2021cross} & 17.94 & 21.44 & 26.17 & 33.33 & 30.91\\
BL+IADM \cite{liu2021cross} & 15.61 & 19.95 & 24.69 & 32.89 & 28.18\\
CmCaF \cite{li2022rgb} & 15.87 & 19.92 & 24.65 & \textbf{28.01} & 29.31\\
\hline
Ours & \textbf{14.32} & \textbf{18.91} & \textbf{23.81} & 32.47 & \textbf{26.01}\\
\hline
\end{tabular}
}
\end{center}
  \vspace{-0.3cm}
\end{table}
\setlength{\tabcolsep}{4pt}

\setlength{\tabcolsep}{4pt}
\begin{table}[t]
\begin{center}
\caption{Counting performance evaluation on ShanghaiTechRGBD.
}
\label{table:4}
\resizebox{0.89\columnwidth}{!}{
\begin{tabular}{cccccc}
\hline
Backbone &  GAME(0)$\downarrow$ & GAME(1)$\downarrow$ & GAME(2)$\downarrow$ & GAME(3)$\downarrow$ & RMSE$\downarrow$\\
\hline
\noalign{\smallskip}
UCNet \cite{zhang2020uc} & 10.81 & 15.24 & 22.04 & 32.98 & 15.70\\
HDFNet \cite{pang2020hierarchical} & 8.32 & 13.93 & 17.97 & 22.62 & 13.01\\
BBSNet \cite{fan2020bbs} & 6.26 & 8.53 & 11.80 & 16.46 & 9.26\\
DetNet \cite{liu2018decidenet} & 9.74 & - & - & - & 13.14\\
CL \cite{idrees2018composition} & 7.32 & - & - & - & 10.48\\
RDNet \cite{sindagi2017generating} & 4.96 & - & - & - & 7.22\\
MCNN+IADM \cite{liu2021cross} & 9.61 & 11.89 & 15.44 & 20.69 & 14.52\\
BL+IADM \cite{liu2021cross} & 7.13 & 9.28 & 13.00 & 19.53 & 10.27\\
CSRNet+IADM \cite{liu2021cross} & \textbf{4.38} & \textbf{5.95} & \textbf{8.02} & \textbf{11.02} & 7.06\\
\hline
Ours & 4.39 & 6.47 & 8.82 & 11.76 & \textbf{6.39} \\
\hline
\end{tabular}
}
\end{center}
\end{table}
\setlength{\tabcolsep}{4pt}

\subsection{Comparison with previous methods}
In addition, to evaluate that our proposed model can be more effective and efficient, we compare our method with state-of-the-art methods on RGB-T and RGB-D crowd counting tasks.
For fair comparisons, we directly use the optimal results of the comparison methods in the original papers.
As shown in Table~\ref{table:3}, compared with multimodal models, including UCNet \cite{zhang2020uc}, HDFNet \cite{pang2020hierarchical}, and BBSNet \cite{zhang2019wide}, our method outperforms these models by a wide margin on all evaluation metrics. Moreover, our model also achieves better performance on the RGBT-CC dataset compared to specially designed crowd counting models, including IADM \cite{liu2021cross}, MVMS \cite{fan2020bbs}, and CmCaF \cite{li2022rgb}. Especially, our CSCA implemented with the BL backbone achieves state-of-the-art performance on almost all evaluation metrics.

For the RGB-D crowd counting task, we also compare two categories of methods including multimodal methods and specially-designed models for cross-modal crowd counting. For fair comparisons,  we use them to estimate the crowd counts by following the work \cite{liu2021cross}. The performance of all comparison methods is shown in Table~\ref{table:4}. The results of the proposed method on the ShanghaiTechRGBD dataset are encouraging, as we used the same parameters as our experiments on the RGBT-CC dataset and did not perform fine-tuning.
Moreover, our model implemented with the CSRNet backbone outperforms most advanced methods and achieves a new state-of-the-art result of 6.39 in RMSE, the main evaluation metric for crowd counting.


\setlength{\tabcolsep}{4pt}
\begin{table}[t]
\begin{center}
\caption{Ablation study using the BL backbone  on RGBT-CC.
}
\label{table:5}
\resizebox{0.89\columnwidth}{!}{
\begin{tabular}{cccccccc}
\hline
Method & GAME(0)$\downarrow$ & GAME(1)$\downarrow$ & GAME(2)$\downarrow$ & GAME(3)$\downarrow$ & RMSE$\downarrow$\\
\hline
\noalign{\smallskip}
RGB & 32.89 & 38.81 & 44.49 & 53.44 & 59.49\\
T & 17.80 & 22.88 & 28.50 & 37.30 & 30.34\\
\hline
Early fusion & 18.25 & 22.35 & 26.78 & 34.96 & 37.71\\
Late fusion & 16.63 & 20.33 & 24.81 & 32.59 & 32.66\\
\hline
NL & 17.91 & 21.12 & 25.06 & 32.61 & 32.60\\
SCA & 17.85 & 21.13 & 25.07 & 32.56 & 30.60\\
CFA & 17.78 & 21.36 & 25.69 & 33.14 & 32.37\\
NL+CFA & 16.69 & 20.75 & 25.63 & 33.92 & 29.39\\
CSCA (SCA+CFA) & \textbf{14.32} & \textbf{18.91} & \textbf{23.81} & \textbf{32.47} & \textbf{26.01}\\
\hline
\end{tabular}
}
\end{center}
\end{table}
\setlength{\tabcolsep}{4pt}

\setlength{\tabcolsep}{4pt}
\begin{table}[t]
\begin{center}
  \vspace{-0.3cm}
\caption{Ablation study using the BL backbone on ShanghaiTechRGBD. Since methods based on non-local blocks run out of memory, we do not report those results.
}
\label{table:6}
\resizebox{0.89\columnwidth}{!}{
\begin{tabular}{cccccccc}
\hline
Method & GAME(0)$\downarrow$ & GAME(1)$\downarrow$ & GAME(2)$\downarrow$ & GAME(3)$\downarrow$ & RMSE$\downarrow$\\
\hline
\noalign{\smallskip}
RGB & 6.88 & 8.62 & 11.86 & 18.31 & 10.44\\
D & 15.17 & 17.92 & 22.59 & 30.76 & 23.65\\
\hline
Early fusion & 7.04 & 8.74 & 11.82 & 18.33 & 10.62\\
Late fusion & 6.55 & 8.01 & 10.45 & 15.43 & 9.93\\
\hline
SCA & 6.94 & 8.19 & 10.48 & 15.91 & 10.61\\
CFA & 6.43 & 8.07 & 10.80 & 16.12 & 9.86\\
CSCA (SCA+CFA) & \textbf{5.68} & \textbf{7.70} & \textbf{10.45} & \textbf{15.88} & \textbf{8.66}\\
\hline

\hline
\end{tabular}
}
\end{center}
\end{table}
  \vspace{-0.3cm}
\setlength{\tabcolsep}{4pt}

\subsection{Ablation studies}
To verify the effectiveness of the proposed universal framework for cross-modal crowd counting, we investigate how different inputs affect the model. 
First, we feed RGB images and thermal (or depth) images into the network, respectively, following the unimodal crowd counting setting of the backbone network. 
Then we adopt the early fusion strategy, feeding the concatenation of RGB and thermal images into BL network.
We also apply late fusion where the different modal features are extracted separately with the backbone network and then combine multimodal features in the last layer to estimate the crowd density map.
From Table~\ref{table:5} and Table~\ref{table:6}, we can observe an interesting finding that thermal images are more discriminative than RGB images for the RGBT-CC dataset.
However, RGB images are more efficient than depth images on ShanghaiTechRGBD. 
Interestingly, naive feature fusion strategies such as early fusion and late fusion even damage the original accuracy.

To explore the effectiveness of each CSCA component, we remove the SCA block and CFA block, respectively. 
As shown in Table~\ref{table:5} and Table~\ref{table:6}, regardless of which component is removed, the performance of our model would drop severely. These results suggest that the main components of CSCA work complementary.
Additionally, we perform additional experiments replacing the SCA block with the typical non-local block.
Remarkably, our CSCA is far superior to non-local-based methods in terms of various evaluation metrics.
Since the non-local-based methods were out of memory on the ShanghaiTechRGBD dataset where image resolution is $1920\times1080$, we could not report the results using non-local blocks.
These results suggest that the SCA blocks have more practical potential than the typical non-local blocks.

To further validate the complementarity of the multimodal data for crowd counting, we infer our model using BL backbone under different illumination conditions on the RGBT-CC dataset. As shown in Table~\ref{table:7}, we conduct experiments under bright and dark scenes, with three different input settings, including unimodal RGB images, unimodal thermal images, and RGB-T multimodal images, respectively. Notably, thermal images can provide more discriminative features than RGB images, especially in lower illumination cases. Moreover, comparing the single-modal results, it is easy to conclude that our CSCA can complementarily fuse thermal information and RGB optical information.

\setlength{\tabcolsep}{2pt}
\begin{table}[t]
\begin{center}
\caption{The performance of our model implemented with BL backbone under different illumination conditions on RGBT-CC.
}
\label{table:7}
\resizebox{0.87\columnwidth}{!}{
\begin{tabular}{ccccccc}
\hline
Illumination & Input & GAME(0)$\downarrow$ & GAME(1)$\downarrow$ & GAME(2)$\downarrow$ & GAME(3)$\downarrow$ & RMSE$\downarrow$\\
\hline
\multirow{3}*{Brightness} & RGB & 26.96 & 31.82 & 37.07 & 46.00 & 48.42\\
~ & T & 19.19 & 24.07 & 30.28 & 39.93 & 32.07 \\
~ & RGB-T & \textbf{14.41} & \textbf{18.85} & \textbf{24.71} & \textbf{34.20} & \textbf{24.74}\\
\hline
\multirow{3}*{Darkness} & RGB & 45.78 & 51.03 & 54.65 & 62.51 & 87.29\\
~ & T & 16.38 & 21.65 & 26.67 & 34.59 & 28.45 \\
~ & RGB-T & \textbf{14.22} & \textbf{18.97} & \textbf{22.89} & \textbf{30.69} & \textbf{27.25}\\
\hline
\end{tabular}
}
\end{center}
\vspace{-0.3cm}
\end{table}
\setlength{\tabcolsep}{2pt}

\section{Qualitative Results}

\begin{figure*}[t]
\centering
 \setlength{\abovecaptionskip}{0.cm}
\vspace{-0.3cm}
\includegraphics[scale=0.6]{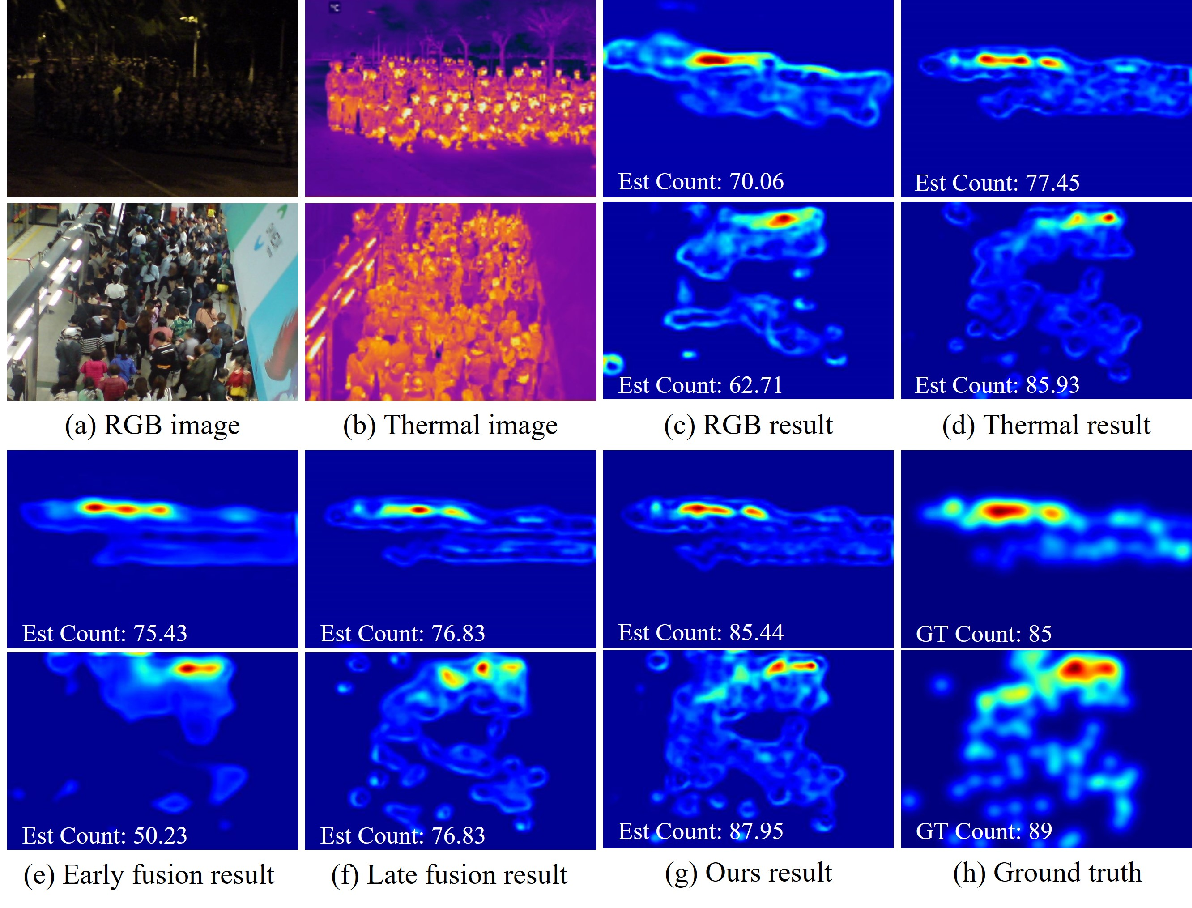} 
\caption{
Qualitative Results. (a) and (b) show the input RGB images
and thermal images in different illumination conditions. (c)-(d) are the results of the RGB-based and thermal-based network. (e)-(f) are the results of the early fusion and late fusion. (g) and (h) refers to the results of our CSCA and the ground truth.
}
\label{Fig:4}
\vspace{-0.5cm}
\end{figure*}

From the visualization results in cases (a) to (d) of Fig.~\ref{Fig:4}, we can easily find that thermal images can facilitate the crowd counting task better than RGB images. Moreover, as shown in Fig.~\ref{Fig:4}-(b), thermal images can also provide additional crowd location information to a certain extent, especially in dark scenes or bright scenes with less light noise.
As we discussed earlier, inappropriate fusions fail to exploit the potential complementarity of multimodal data and even degrade the performance, such as the early fusion and late fusion shown in Fig.~\ref{Fig:4}-(e) and Fig.~\ref{Fig:4}-(f). Our proposed CSCA, a plug-and-play module, can achieve significant improvements for cross-modal crowd counting by simply integrating into the backbone network as shown in Fig.~\ref{Fig:4}-(g).
This result shows the effectiveness of CSCA's complementary multi-modal fusion.

\section{Conclusion}
Whereas previous crowd counting approaches have mainly focused on RGB-based frameworks, we focus on a cross-modal representation that can additionally utilize thermal or depth images. Specifically, we propose a  novel plug-and-play module named Cross-modal Spatio-Channel Attention (CSCA), which can be integrated with any RGB-based model architecture.
Our CSCA blocks consist of two main modules. First, SCA is designed to spatially capture global feature correlations among multimodal data.
The proposed SCA to deal with cross-modal relationships is computationally more efficient than existing non-local blocks by re-assembling the feature maps at the spatial level.
Furthermore,  our CFA module adaptively recalibrates the spatially-correlated features at the channel level.
Extensive experiments on the RGBT-CC and ShanghaiTechRGBD datasets show the effectiveness and superiority of our method for multimodal crowd counting.
Additionally, we conduct extensive comparative experiments in various scenarios using unimodal or multimodal representations, which have not been well addressed in crowd counting. 
These results are expected to be used as baselines for subsequent studies for cross-modal crowd counting.
\\

\noindent \textbf{Acknowledgements.} This work was partly supported by
Samsung Research Funding \& Incubation Center of Samsung Electronics under Project Number SRFC-TD2103-01 and INHA UNIVERSITY Research Grant.

\bibliographystyle{splncs04}
\bibliography{mybibliography}
\end{document}